\documentclass[]{article}
\usepackage{graphicx,color}

\title{Exoway: an exoskeleton on actuated wheels}
\author{D. Abruzzese, D. Carnevale, A. Monti,  C. Possieri, S. Rossi, \\
M. Sassano, P. P. Valentini}

\begin{document}

\maketitle

\begin{abstract}
In this short work we present a low cost
exoskeleton with actuated wheels that allows movements 
as well as skating-like steps. 
The simple structure and the actuated wheels allows to 
minimize the use of motors for locomotion.
The structure is stabilized by an active control system
that balances the structure and permit to be maneuvered
by the driver whose commands are acquired by a dedicated 
hardware interface. 
\end{abstract}

\section{Introduction}
\subsection{General Target}
The main objective of the EXOWAY project consists in designing and prototyping an advanced self-balanced EXOskeleton with actuated wheels, similarly to segWAY, simultaneously capable of high-velocity
movement on flat surfaces and of climbing stairs/steps to significantly increase  mobility of people with impaired lower limbs motor functions.

\subsection{State of the Art}
Approximately 250 to 500 thousands people suffer from Spinal Cord Injury (SCI) every day, representing a global incidence of 23 cases per million \cite{1}. The most common causes of SCI are traffic accidents,
gunshot injuries, knife injuries, falls and sports injuries \cite{2}, which cause a lesion of the neural components of the spinal cord, such as medulla, medullary cone and/or cauda equina \cite{3}. In SCI patients
muscular recruitment and motor planning are compromised; thus, gait impairment represents one of the most critical disabilities and may cause the complete loss of voluntary control of the leg muscles \cite{4}. In
addition, patients with SCI are subjected to secondary complications \cite{7}, such as psychological, cutaneous and cardiorespiratory problems \cite{5}. Moreover, the number of people that have limited capability of
autonomous gait increase impressively if we include stroke and Parkinson’s disease patients. The overall picture is expected to become even more critical in the very near future due to rapidly aging
population.
Canes, walkers or wheelchairs are useful devices that permit a certain level of mobility \cite{8}; however, these traditional assistive devices do not allow patients to achieve a full social participation, since they limit
their ability to perform daily activities and - especially concerning the case of wheelchairs - they can lead to muscle atrophy due to prolonged usage and do not provide a sufficient stimulation to the blood
circulatory system.
Innovative mobility devices have been developed in recent years, encompassing powered lower limb exoskeletons \cite{8} and powered wheelchairs \cite{9}. Among the former, ReWalkTM \cite{10} is the most widespread
in both domestic and clinical contexts. ReWalk is an assistive device that provides powered hip and knee motion, allowing patients with SCI to stand upright, walk, turn and ascend/descend stairs. Each step
is generated by means of the measurement of user’s trunk movements gathered by a tilt sensor. The major disadvantage is that the patient needs to use canes to stabilize the walk. This aspect requires
patient training and use of upper limbs to permit locomotion, precluding to such patients all daily life/work activities that require the use of upper limbs. An additional limitation, shared also by similar devices
(Indego, EksoGT, ExoAtlet, ecc..), is the limited (maximum) achievable walking speed, that is critically low for the first self-balancing powered exoskeleton commercially available, namely the REX® \cite{11}. REX,
equipped with motors and hardware dedicated to ensure self balanced walking, results in an heavy and bulky structure whose price overpasses alternative exoskeletons by more than 40k$ (approx. price of
exoskeletons is 70k$).

\section{Mechanical design}

To address and circumvent the severe limitations discussed above, the EXOWAY is envisioned to suitably combine the features of a segway and of an exoskeleton (Figure \ref{figure1}).

\begin{figure}[h!]
      \centering
      \resizebox{0.8\columnwidth}{!}{\includegraphics{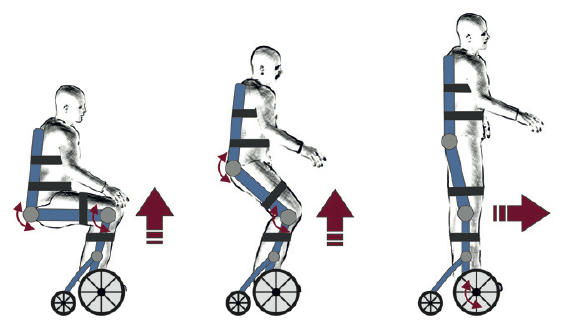}}
      \caption{Sketch of the EXOWAY mechanical structure and 
       its motion with paired wheels.}
      \label{figure1}
\end{figure}

The two independent actuated wheels at the base of the structure allow to simultaneously obtain faster locomotion while lightening the overall architecture. Two additional passive rear wheels can be  linked at
each of the exoskeleton’s shins by means of (active or passive) rotational 
joints to provide a wider support on the ground, thus reducing the stabilization effort required by the active
wheels, increasing patient’s comfort, and improving energy consumption. 
By potentially acting on the rotational joint of such passive wheels their configuration can be modified to provide additional stabilizing
contact points when climbing and descending stairs or small steps as 
sketched in Figure \ref{figure2}.

\begin{figure}[h!]
      \centering
      \resizebox{0.4\columnwidth}{!}{\includegraphics{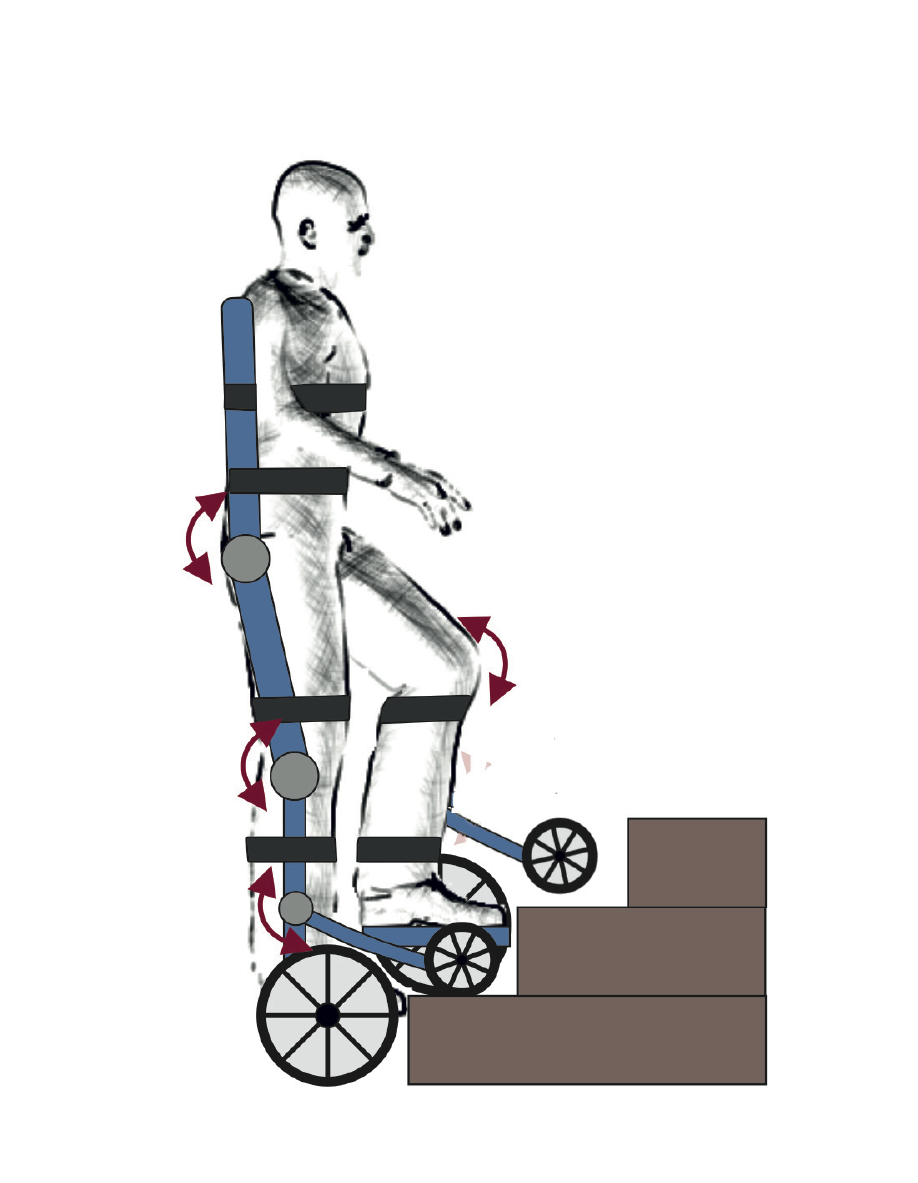}}
      \caption{The EXOWAY climbing stairs.}
      \label{figure1}
\end{figure}

The optimal mechanical design of such passive wheels allows 
to improve stability and reduce encumbrance.
The envisioned design is highly promising in its functionalities and, 
to the best of our knowledge, completely innovative with respect to existing solutions given that the two legs are independent an not fixed as in
Segways and then also a \textbf{skating-like} gait is allowed resulting
into an increased stability during locomotion and possible 
blood circulatory benefits.\\

Motors can be added at each exoskeleton’s knee and hip joint to provide basic movements such us stand-up/down and climb stairs or the rest of the structure 
can be simply made with passive (dissipative) joints. 
The overall structure is composed by seven frames rigidly (passive dampers
can be introduced to reduce stiffness) connected to trunk, thighs, shins and feet. The envisioned structure allows to mimic what we can define a “soft-gait”, i.e. coordinated movements between the two
wheeled legs resembling a straight skating gait: this can be achieved controlling relative speed between the two wheels, knee and hip motors. 

The small rear wheels can be even detached from the EXOWAY 
structure when using it in planar and constrained spaces at the cost
of an higher energy consumption for the stabilization.

In case the battery or the control system fail, a system of passive 
elements is installed to open wide the legs and guarantee a stable 
position.

\section{Electronics and control}

The Exoway is provided with a self-balancing capability thanks to a 
control board that exploits IMUs,  tilting sensors, optical 
encoders and force sensors as well as stereo camera and ranging sensors
to estimate  the structure tilt and position (VSLAM) in the surrounding space
also in order to avoid collisions. Indeed the control system does provide
also to collision avoidance within a dynamic environment.\\
The patitent/user can select direction and speeds by mean of 
a sensorized glow, voice commands, and joysticks.

\end{document}